\documentclass{article}

\usepackage{microtype}
\usepackage{graphicx}
\usepackage{subfigure}
\usepackage{booktabs} 
\usepackage[table]{xcolor}
\usepackage[export]{adjustbox}

\usepackage{hyperref}
\usepackage{wrapfig}

\usepackage[accepted]{icml2024}

\usepackage{amsmath}
\usepackage{amssymb}
\usepackage{mathtools}
\usepackage{amsthm}

\usepackage[capitalize,noabbrev]{cleveref}

\theoremstyle{plain}
\newtheorem{theorem}{Theorem}[section]
\newtheorem{proposition}[theorem]{Proposition}
\newtheorem{lemma}[theorem]{Lemma}

\theoremstyle{definition}
\newtheorem{definition}[theorem]{Definition}

\theoremstyle{remark}
\newtheorem{remark}[theorem]{Remark}

\usepackage[textsize=tiny]{todonotes}

\icmltitlerunning{Message Detouring: A Simple Yet Effective Cycle Representation for Expressive Graph Learning}

\begin{document}
\twocolumn[
\icmltitle{Message Detouring: A Simple Yet Effective Cycle Representation \\for Expressive Graph Learning
}




\begin{icmlauthorlist}
    \vspace{-1em}
\icmlauthor{Ziquan Wei}{yyy,equal}
\icmlauthor{Tingting Dan}{equal}
\icmlauthor{Guorong Wu}{equal,yyy}

$^{\text{1}}$ Department of Computer Science, University of North Carolina at Chapel Hill, Chapel Hill, USA

$^{\text{2}}$ Department of Psychiatry, University of North Carolina at Chapel Hill, Chapel Hill, USA

\texttt{{guorong\_wu@med.unc.edu}, \url{https://anonymous.4open.science/r/message-detouring-neural-network-65A8}}
\end{icmlauthorlist}

\icmlaffiliation{yyy}{Department of Computer Science, University of North Carolina at Chapel Hill, Chapel Hill, USA}
\icmlaffiliation{equal}{Department of Psychiatry, University of North Carolina at Chapel Hill, Chapel Hill, USA}

\icmlcorrespondingauthor{Guorong Wu}{guorong\_wu@med.unc.edu}

\icmlkeywords{Edge detour, Detour path, Message Detouring, Message Passing, Graph expressive power, Graph isomorphic test, Graph learning, Graph classification, Node classification}

\vskip 0.3in
]




\begin{abstract}

Graph learning is crucial in the fields of bioinformatics, social networks, and chemicals. Although high-order graphlets, such as cycles, are critical to achieving an informative graph representation for node classification, edge prediction, and graph recognition, modeling high-order topological characteristics poses significant computational challenges, restricting its widespread applications in machine learning. To address this limitation, we introduce the concept of \textit{message detouring} to hierarchically characterize cycle representation throughout the entire graph, which capitalizes on the contrast between the shortest and longest pathways within a range of local topologies associated with each graph node. The topological feature representations derived from our message detouring landscape demonstrate comparable expressive power to high-order \textit{Weisfeiler-Lehman} (WL) tests but much less computational demands. In addition to the integration with graph kernel and message passing neural networks, we present a novel message detouring neural network, which uses Transformer backbone to integrate cycle representations across nodes and edges. Aside from theoretical results, experimental results on expressiveness, graph classification, and node classification show message detouring can significantly outperform current counterpart approaches on various benchmark datasets.


\end{abstract}

\section{Introduction}
\label{intro}

%



Meaning feature representations for graph substructures (also referred to by various names including graphlets, motifs, subgraphs, and graph fragments) have been extensively investigated \cite{murray2009rise,jiang2010finding,koyuturk2004efficient}, with widespread applications in reasoning path \cite{ding2019cognitive} and cycle basis \cite{yan2022cycle} for knowledge graph, as well as neural fingerprint \cite{duvenaud2015convolutional}, junction tree autoencoder \cite{jin2018junction} and Cellular Weisfeiler Lehman (CWL) testing \cite{bodnar2021weisfeiler} for molecular substructure encoding. To practically encode such substructures, high-order approaches are proposed for learning-based methods by building hyper-graphs \cite{gao2022hgnn+} or node tuples \cite{zhao2022practical}, such as $k$-IGN \cite{chen2020can} and message passing simplicity neural network (MPSN) \cite{bodnar2021weisfeilerb}. Although increasing evidence suggests that these neural networks have the potential to express graph structures beyond node and edge, the computational cost is often demanding.

\begin{figure}[t]
    \centering
    \includegraphics[width=.5\textwidth]{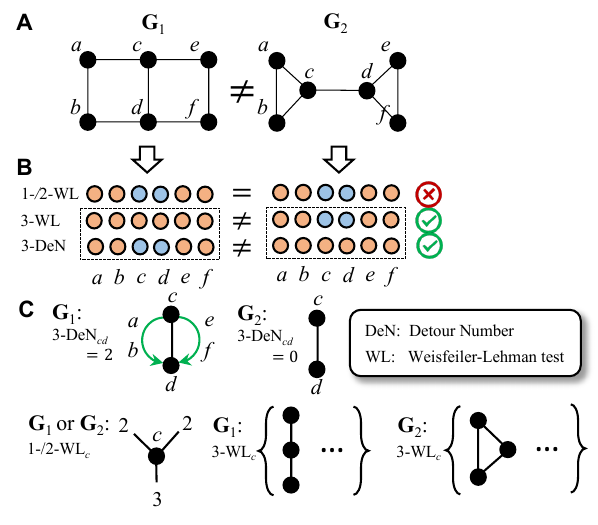}
    \vspace{-3em}
    \caption{Comparison between 1-WL initialized by our node-wise cycle representation 3-DeN (see details in Section \ref{DeN} Detour number) and node degree initialized 1-, 2-, or 3-WL tests. (A): An example of two non-isomorphic graphs $\mathbf{G}_1$ and $\mathbf{G}_2$. (B): Final converged coloring results of various WL tests. (C): The 3-DeN shows enough expressive power to differentiate $\mathbf{G}_1$ that has 2 detour paths (green arrows) and $\mathbf{G}_2$ that has none with respect to edge ${cd}$. In contrast, conventional WL tests fail until using 3-tuples of nodes where 3-WL test is much more computationally demanding than our detouring mechanism.}
    \label{fig:den_example}
    \vspace{-1.5em}
\end{figure}

The advancement of graph isomorphic tests has been extended to message passing neural networks (MPNN) based on group theories \cite{xu2018powerful,bronstein2017geometric,atz2021geometric}. However, these methods face challenges associated with the limitation of message passing, including issues like over-smoothing \cite{keriven2022not}, over-squashing \cite{akansha2023over}, and limited expressive power compared to the first order Weisfeiler-Lehman test (1-WL) \cite{balcilar2021breaking,balcilar2021analyzing}. More critically, MPNN has been proved only capable of expressing \textit{star-shaped} topology structures \cite{arvind2020weisfeiler,chen2020can}.
Herein, message passing mechanism solely through the shortest path fails to adequately capture the \textit{cycle} structures \cite{huang2022boosting}. As illustrated in Fig. \ref{fig:den_example} (A) and (B), $\mathbf{G}_1, \mathbf{G}_2$ are 1- or 2-WL equivalent non-isomorphic graphs. The reason for the failure of 1-/2-WL tests is that nodes $\{c, d\}$ share the same star-shaped structures between $\mathbf G_1$ and $\mathbf G_2$ shown in the left corner of (C), while the number of cycles associated with nodes $c$, $d$ is completely different: Two (\#1: \textit{$\overrightarrow{cabdc}$} and \#2: \textit{$\overrightarrow{cefdc}$}) for each node in $\mathbf G_1$, but only one in $\mathbf G_2$. As a result, the colorings of two graphs by 1-/2-WL are indistinguishable (1$^{st}$ row in (B)) but distinguishable for 3-order WL (3-WL, 3$^{rd}$ row in (B)). 

It is worth noting that nodes involving with multiple cycles can always be divided into (1) node set associated with the shortest path and (2) the remaining nodes along the additional paths that makes up the cycle (green arrows in(C)). In this context, we refer the additional paths as \textit{detour path} (see \textbf{Definition \ref{intro_def1}}) in this work, and essentially, they are equivalent to a cycle. To this end, by allowing up to three nodes in the detour path, the \textit{detour number} ($3$-DeN) emerges as a putative cycle representation at both node and edge level. Compared to $2$-DeN (no greater than 2 nodes in the detour), $3$-DeN manifests enough expressive power in recognizing the topological difference on edge $cd$ as depicted in the last row in (B).


\begin{definition}[Detour path]
\label{intro_def1}
    Suppose $\mathcal P_1$ and $\mathcal P_2$ are two paths in the graph. If $\exists \mathcal P_1,\mathcal P_2$ connecting the same endpoints and $\mathcal P_1\neq\mathcal P_2,|\mathcal P_1| \leq |\mathcal P_2|$, then $\mathcal P_1\cup\mathcal P_2$ is a cycle and $\mathcal P_2$ is a detour path with respect to the path $\mathcal P_1$.
    \vspace{-0.5em}
\end{definition}

Since \cite{chartrand1993detour} framed the detour distance of the graph, the detour pattern has been researched in graph theory \cite{santhakumaran2010edge} and discrete mathmetics \cite{parthipan2020isolated} for years. \cite{vaidya2019detour} proposed a concept of detour domination theory by replacing the usual distance (alternative name of the shortest distance) with detour distance. As a result, the dominating importance of detour inspires us with its potential for expressive graph learning. To the best of our knowledge, no existing work in the realm of machine learning has yet explored the concept of detours as described here. In light of this, we seek to bridge the gap between notion of detour path in graph theory and graph data learning. 

Beyond the group theory, network detours are fundamentals in neuroscience and real-world networks, respectively. Brain functions utilize a detour path along with the shortest path to amplify functional connectivity \cite{goni2014resting}. Detour route analysis in traffic tasks \cite{feng2022understanding}, tourism detour spots \cite{hirota2019generating}, and third-person (referring to a detour connection) effects on social-media \cite{antonopoulos2015web} are real-world instances of network detour patterns contributing to practical problems. In this context, a putative cycle representation is appealing for expressive graph learning on these network data using the idea of detour. 

In this work, we propose a principled framework of message detouring, anchored on the topological characteristics of detour path which emerges as a simple yet effective cycle representation both at node and edge level for expressive graph learning. Furthermore, we have derived theoretical proof that the new message detouring mechanism is as powerful as high-order WL tests yet demands much fewer computational resources. From the perspective of applicability, we demonstrate that it is straightforward to plug in our detouring mechanism to current graph-based models, such as graph kernel and graph neural networks, and instantly enhance the graph learning performance. Additionally, we take a step further by devising a Transformer-style deep model, coined message detouring neural network (MDNN), 
where the attention mechanism allows featurizing nodes with detour information. 
We have comprehensively evaluated the expressive power, node classification, and graph classification on both synthetic and real-world graph datasets, where our method (steered by message detouring mechanism) achieves better performance compared to message passing counterparts.

We briefly summarize our major contributions as follows. 
{(1)} A simple yet effective message detouring mechanism to express high-order cyclic structure in the graph at both node and edge level.
{(2)} A collection of theoretical results of analyzing the expressive power of detouring mechanism via isomorphic test and cycle counting ability.
{(3)} A scalable framework to integrate message detouring mechanism into graph data learning with several new deep models showcasing SOTA performance on both synthetic and real-world graph datasets. 

\section{Preliminaries}

Graphs discussed in this work are undirected and finite. We assume a graph is denoted by $G=\{V, E\}$, where $V$ is the set of vertices/nodes and $E$ is the set of edges. If we assume the cardinality of a set is denoted by $|\cdot|$, then the degree of the node $V_i$ is $D(V_i) = |\mathcal N(V_i)|$, where $\mathcal N(V_i)$ is the set of neighborhoods of $V_i$. Multiset is denoted by $\{\!\!\{\cdot\}\!\!\}$, which allows duplicate elements.  The element of a set/multiset is noted by the name of the set/multiset with subscripts.

\subsection{What is a detour path?}

In graph theory, the detour distance proposed by \cite{chartrand1993detour} denotes the longest path between two endpoints. The concept of the detour, however, has not been brought into the field of deep geometric learning since its presence. The main reason is the nature of NP hard to find the longest path. Instead of the longest path that could be found in $O(|V|!)$ time cost to connect any pair of nodes in a complete graph, longer paths connecting neighborhoods are efficient and powerful approximation in representing graph topology. Thus, longer paths corresponding to an edge, i.e., detour path set in \textbf{Definition \ref{def1}}, are the cornerstone for message detouring.
\begin{definition}[Detour path set]\label{def1}
\textit{Given two nodes $V_i$ and $V_j$ on a graph $G$, if $\mathcal P_{ij}$ is a path that has endpoints $V_i,V_j$, then the detour path set is defined as $\Omega^k_{ij}\triangleq\{\mathcal P_{ij}|\ 1<|\mathcal P_{ij}|\leq k\}$.}
\end{definition}


Finding all detour path sets of a graph is running a modified Depth-First Search (DFS) \cite{sedgewick2001algorithms}. Searching stops at the depth $k$. A single path can be found in $O(V+E)$ time.
In comparison with $k$-WL tests, finding a detour path set is cheaper than $k$-WL if $k>2$, which strictly costs $O(|V|^k)$ computational time regardless of the edge number. 

\subsection{Weisfeiler-Leman tests}

\begin{figure}[t]
    \centering
    \vspace{-.7em}
    \includegraphics[width=.5\textwidth]{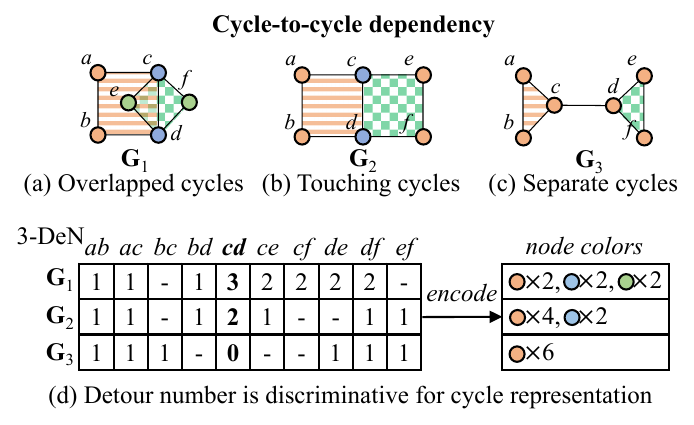}
    \vspace{-2.5em}
    \caption{The presence of cycles poses challenges to recognize isomorphism via message passing mechanism. Graphs with (a) overlapped, (b) touching, and (c) separate cycles are 1-WL equivalent when node degree is identical. (d) Topological differences between graphs can be captured by the number of detour paths, i.e. detour number (DeN). For example, the pattern of DeN on edge $cd$ is well aligned with the trending of cycle-to-cycle relationship from (a)-(c), i.e., DeN on edge $cd$ consistently decrease as the topological context between two underlying cycles (filled with different strip patterns) evolves from highly overlapped to no overlap at all. The sketch of detour paths in $\mathbf{G}_1$-$\mathbf{G}_3$ is shown in Appendix \ref{appendix_detour_path}.}
    \label{fig:wl_failure}
\vspace{-1.0em}    
\end{figure}

The Weisfeiler-Leman (WL) graph isomorphism test \cite{leman1968reduction} is an iterative algorithm by refining the initial state of a node $V_i$ with any hash method until convergence \cite{morris2021weisfeiler}:
\begin{equation}\label{wltest}
    wl^{l+1}(V_i) = \text{hash}(wl^l(V_i), \{\!\!\{wl^l(V_j)|j\in\mathcal N(V_i)\}\!\!\}),
\end{equation}
where $l$ is the iteration number. In a nutshell, $\text{hash}(\cdot)$ generates distinct color patterns based on topological measurements associated with each node (such as node-wise connectivity degree). A typical example of graph coloring is shown in Fig. \ref{fig:wl_failure} top. $k$-WL tests extend the original WL test by allowing the input to be a tuple of $k$ nodes, namely, $wl^l(\{V^k_i\})$. $k$-WL tests capture more information about the node community and are capable of distinguishing graphs that are $(k-1)$-WL equivalent, yet at the expense of computational cost iterating $\frac{|V|!}{|k|!}$ tuples. Such high-order method is introduced to the field of graph learning to gain higher expressive power \cite{maron2019provably,maron2018invariant,maron2019universality}, with a family of universal $k$-order graph networks and stated a theory of the most powerful MPNN comparing to $k$-WL tests when $k$ is greater than two.

Specifically, the initial state in WL tests is based on the message passing number, i.e., the node degree of connectivity. Related works \cite{azizian2020expressive,oono2019graph,huang2022boosting,yan2022cycle} claim that the investigation on expressive power of a graph neural network should take the ability of substructure detection, e.g. cycle into account, in addition to graph isomorphic tests. However, observations and theoretical analyses \cite{xu2018powerful,chen2020can} have revealed the limitations of 1-WL for the cycle, and hence MPNN as well.  

\subsection{An in-depth understanding of challenges for expressing cycles in MPNN}\label{why_detour}

The cycle in a graph is the barrier of message passing and aggregation algorithm for distinguishing non-isomorphic graphs. Node-wise message passing, e.g. graph convolution, has been proved unable to express cycle \cite{huang2022boosting}. The example in Fig. \ref{fig:den_example} shows that $k$-WL is powerful enough to express isomorphism only if $k>2$. In fact, node-wise differences on node $c$ and $d$ between graphs $\mathbf{G}_1$ and $\mathbf{G}_2$ in Fig. \ref{fig:den_example} can be described by the contrast of detour paths denoted by green arrows as shown in Fig. \ref{fig:den_example} (C).

\begin{remark}\label{1wl2wl}
\textit{For graphs with unweighted edges, 1-WL and 2-WL are known to have the same discriminative power \cite{maron2019provably}.}
\end{remark}

\begin{remark}\label{passing_lb}
\textit{1-WL test has no power to assign different states to non-isomorphic graphs, if their cycles have varying counts.
 \cite{arvind2020weisfeiler}.}
\end{remark}
Together, Remark \ref{1wl2wl} and Remark \ref{passing_lb} lead the fact that 1- or 2-WL cannot theoretically distinguish non-isomorphic graphs since cycle counts vary. In Fig. \ref{fig:wl_failure}, $\mathbf{G}_1$ in (a) has 6 cycles, $\mathbf{G}_2$ in (b) has 3 cycles, and $\mathbf{G}_3$ in (c) has 2 cycles. In addition to the varying number of cycles, there are diverse topological relationship between cycles in the graph. For example, two cycles may (a) overlap, (b) touches, and (c) be far apart in the graph. In the context of WL tests, $\mathbf{G}_2$ and $\mathbf{G}_3$ remain isomorphic until the topology manifests difference using 3-path-long cycles.

\section{Theoretical results}\label{sec:exp_pow}
To address the challenges of expressing cycles, we devise a simple yet effective cycle representations using the notion of detour path. Here, we provide the theoretical proof that characterizing detour paths edge by edge (an efficient approach) is equivalent to modeling the topological representations of cycles (which is computational expensive). 

\subsection{Detour number (DeN)}\label{DeN}

Inspired by the message passing number, we propose the detour number by counting detour paths.

\begin{definition}[Detour number]\label{def2}
\textit{The detour number on edge $E_{ij}$ is defined as the cardinality of a detour path set $\Omega^k_{ij}$ (Definition \ref{def1}), i.e. $k\text{-}DeN_{ij}\triangleq|\Omega^k_{ij}|$.}
\end{definition}
According to Definition \ref{def1}, the union of the detour path and the corresponding shortest path $\mathcal P_{ij}\cup \{E_{ij}\}$ form a cycle given nodes $V_i$ and $V_j$. In this regard, the detour number has the nature of expressing cycles. In Appendix \ref{appendix_detour_path}, we display all possible detour paths for each node on graphs $\mathbf{G}_1$, $\mathbf{G}_2$, and $\mathbf{G}_3$ in Fig. \ref{fig:wl_failure}. In this example, the major difference from $\mathbf{G}_1$ to $\mathbf{G}_3$ is manifested on the edge $E_{cd}$, which is associated with various cycle-to-cycle relationships such as overlapping (a), touching (b), and being far apart (c). As the table shown in Fig. \ref{fig:wl_failure}(d), the detour number 3-DeN (detour length is restricted to no great than three paths) is already able to effectively characterize the topology difference in $\mathbf{G}_1$-$\mathbf{G}_3$, which approaches the similar expressive power of 3-WL test. However, the computational cost of searching detour paths in a $k$-hop graph neighborhood is lower than that of 3-WL test.


\textbf{From edge to node.}
Since the definition of $k\text{-}DeN_{ij}$ is associated with nodes $V_i$ and $V_j$, it is straightforward to encode edge-wise heuristics of detour number into a node-wise feature representation $\Phi_i^k$ for expressing cycles by (1) computing the total degree of the edge-wise detour number for all links connected to the underlying node $V_i$, (2) use hash function in Eq. \ref{wltest} to generate hash value. Specifically, the calculation of $\Phi_i^k$ is formulated as $\Phi^{k}_i=\sum_{j\in\mathcal N(V_i)}k\text{-}DeN_{ij}, \forall k>1$.

In Fig. \ref{fig:wl_failure}(d) right, the color patterns indicate the distinct hash values for each $\Phi_i^3$ in graphs $\mathbf{G}_1$-$\mathbf{G}_3$, where we further color the corresponding nodes for each graph on the top of Fig. \ref{fig:wl_failure}. To that end, at the graph level, the combination of node-wise detour features is distinguishable to separate $\mathbf{G}_1$, $\mathbf{G}_2$, and $\mathbf{G}_3$. 

\subsection{Expressive power}

The expressive power is often benchmarked by the comparison with WL tests and the ability to count substructures for specific problems. In this section, we propose that cycle counting by the detour number, and message detouring is more expressive than conventional message passing in MPNN.


\begin{proposition}\label{den_cycle_counting}
\textit{$\Phi^{k}_i$ equals two times of the cardinality of the set of cycles with length no longer than $(k+1)$ where each cycle includes node $V_i$.}
\end{proposition}

Proof of Proposition \ref{den_cycle_counting} refers to Appendix \ref{proof_den_cycle_counting}. Regarding WL tests, the following theorem states that using detouring, even the node-wise 1-WL test can be as expressive as high-dimensional WL tests.




\begin{theorem}\label{detouring_pw}
\textit{Suppose we use the combination of node degree and the corresponding node-specific detour path, i.e., $\{\!\!\{D(V_i)\}\!\!\}\cup\{\!\!\{k\text{-}DeN_{ij}|j\in\mathcal N(V_i)\}\!\!\}$ (denoted by $\{\!\!\{D,k\text{-}DeN\}\!\!\}$ in brief), to initialize $1$-WL. The express power of such $1$-WL testis higher in expressing isomorphism of connected graphs than the conventional $1$-WL initialized using node degree only. Furthermore, it can reach the expressive power of $k$-WL.}
\end{theorem}

Detailed proof can be found in Appendix \ref{proof_pw}.

\section{Implementations of message detouring }
Cycle is a common topological structure in many graph data such as real-world networks and chemical structures. In this context, the output of message detouring is expected to yield a new message of how an edge is distinct to edges that involve in various cycles, either amplifying signals imitating real-world networks or denoting specific chemical structures. In this section, we translate the theory of detour path for cycle representation to a set of novel graph learning methods by presenting an integrated framework called message detouring. 

\textbf{WL kernels initialized with detour path} 
WL kernels are proposed by directly fitting WL labels into graph learning by considering them as node features. Therefore, with a minimum effort, the $\{\!\!\{D,k$-DeN$\}\!\!\}$-initialized WL labels can utilize the proposed node-wise cycle representation on practical tasks, while original kernels have only the ability for \textit{star-shaped} substructures. We call this kernel WL DeN.

\textbf{DeN-weighted MPNN} 
Graph convolutions essentially perform information exchange by message passing and aggregation with learnable parameters. Similarly, converting one detour path as a new direct (i.e., immediate) connection by weighting the original edges with $k$-DeN can encourage current MPNN to learn the feature of the detour pattern. We call this implementation a DeN-weighted MPNN.

In practice, we concatenate the feature of a graph that has DeN-weighted edges with the feature of the original graph as the output of a layer. As a result, DeN-weighted MPNN works the same as the WL DeN kernel by harnessing both (low-order) star-shaped and (high-order) cyclic structural information.

\begin{figure}[t]
    \centering
    \includegraphics[width=.5\textwidth]{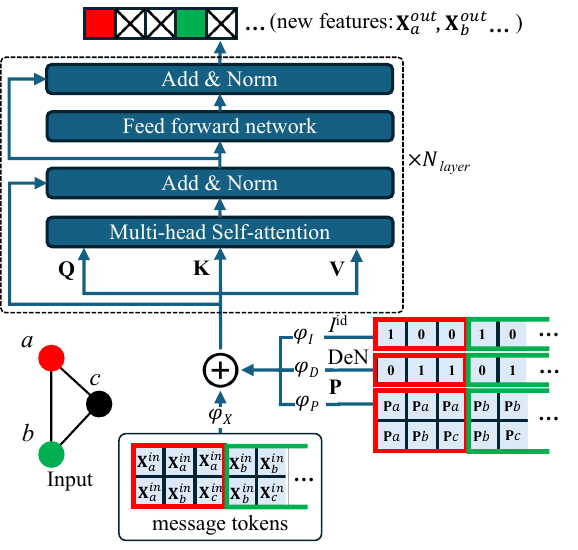}
    \vspace{-2.5em}
    \caption{The flowchart of message detouring neural network (MDNN) with a 3-vertex graph as an example, where node and corresponding tokens framed by the same color. Norm denotes the batch normalization.}
    \vspace{-1em}
    \label{fig:mdnn_flowchart}
\end{figure}

\textbf{Message detouring neural network (MDNN)}
Inspired by recent success of graph Transformers \cite{yun2019graph,kim2022pure}, we propose a new design of Transformer-style GNN named message detouring neural network (MDNN). Specifically, the input to MDNN consists a set of node-wise feature representations, denoted by $\{\mathbf{X}_i^{in} |i=1,...,|V|\}$. The backbone of our MDNN is a Transformer of message detouring which generate new node features $\{\mathbf{X}_i^{out} |i=1,...,|V|\}$ by aggregating detour path and graph spectral information in a multi-head self-attention mechanism \cite{vaswani2017attention}. 

The network architecture of MDNN is shown in Fig. \ref{fig:mdnn_flowchart}. \textit{First}, we generate message token $\mathbf{MSG}_{ij}$ between node $V_i$ and $V_j$ by:
\begin{align}
\begin{split}
    \mathbf{MSG}_{ij} = \varphi_{X}\left(\mathbf{X}^{in}_{i}, \mathbf{X}^{in}_{j}\right) + \varphi_{I}\left({I}^{\text{id}}_{ij}\right)&\\ + \varphi_{P}\left(\mathbf{P}_i,\mathbf{P}_j\right) + \varphi_{D}\left(k\text{-DeN}_{ij}\right)&.
\end{split}
\label{eq.msg}
\end{align}
Specifically, $\varphi_{X}$ is a multi-layer perceptron (MLP) trained to project a pair of input node features $\mathbf{X}^{in}_i$ and $\mathbf{X}^{in}_j$ to a latent $d$-dimensional vector space. As shown in Fig. \ref{fig:mdnn_flowchart} bottom, we concatenate $\mathbf{X}^{in}_i$ with its duplicate if $i=j$. 
Meanwhile, we deploy three MLPs ($\varphi_I$, $\varphi_P$, and $\varphi_D$) to learn the feature representation from node-specific self-loop identifier ${I}^{\text{id}}_{ij}$, eigen-embeddings $\mathbf{P}_i$ and $\mathbf{P}_j$ of graph Laplacian, and detour number $k\text{-DeN}_{ij}$, respectively. Since we allow self-loop in Eq. \ref{eq.msg}, we introduce the token identifier ${I}^{\text{id}}_{ij}$ to indicate whether $V_i$ and $V_j$ form a self-loop, i.e., $I^{\text{id}}_{ij}=1$ if $i=j$ and $0$ otherwise. $\mathbf{P}_i$ is the positional encoding which corresponds to the $i^{th}$ row of eigenmatrix of graph Laplacian (each column is a eigenvector). 


After that, we collect the message token $\mathbf{MSG}_{ij}$ and further learn high-level feature representations using multi-head self-attention as:
\begin{equation}
    \mathbf{X}_{i}^{out} = \mathcal T(\{\mathbf{MSG}_{ij}|j\in\mathcal N(V_i)\cup\{{i}\}\}),
\end{equation}
where the workflow of $\mathcal T(\cdot)=softmax(\frac{QK^T}{\sqrt{d}})V$ is shown in the dashed block in Fig. \ref{fig:mdnn_flowchart}, which each message token $\mathbf{MSG}_{ij}$ is used as query (Q), key (K), and value (V). With no doubt, we can progressively find best feature representation $\mathbf{X}^{out}$ by stacking this Transformer $N_{layer}$ times. 

Since $\varphi_{P}$ is derived by graph Laplacian
and $\varphi_{D}$ is derived by $k\text{-DeN}$, our MDNN is the learnable version of the combination $\{\!\!\{D, k\text{-DeN}\}\!\!\}$.
Afterward, to obtain the final prediction, the new feature of nodes can be trained with various GNNs end-to-end. Detailed implementations refer to Appendix \ref{appendix_imp}. 










\section{Experimental results}

To validate our theoretical results and evaluate the performance of message detouring, our experiments are designed to answer the following questions: \textbf{Q1:} Is the detour number ($k$-DeN) capable of distinguishing more non-isomorphic graph pairs than the original 1- or 2-WL? \textbf{Q2:} To what extent does the performance of real-world graph predictions improve with message detouring compared to traditional message passing? \textbf{Q3:} As a node-wise representation, can the message detouring outperform message passing in node predictions?

\textbf{Datasets} 
To show the expressive power of message detouring, we run the experiments following \cite{kriege2022weisfeiler} by using two synthetic datasets that are designed to be indistinguishable by 1- or 2-WL test, and five real-world datasets of small molecules, bioinformatics, and social networks as well. 
EXP and CEXP are two synthetic datasets from \cite{abboud2020surprising}. They both consist of 600 pairs of non-isomorphic graphs, where 100\% and 50\% of graph pairs are strictly indistinguishable for 1- or 2-WL but distinguishable for 3-WL, respectively. Among the real-world datasets, MUTAG, NCI1, and PTC\_FM are small molecules. ENZYMES and DD are bioinformatics. IMDB-Binary consists of social networks. Chameleon and Squirrel are knowledge graphs from Wikipedia. Actor is a film network from Wikipedia. Cornell, Wisconsin, and Texas are website networks. Small molecules, bioinformatics, and social networks are used for graph classifications, others are for node classifications. We follow the data splits of \cite{errica2019fair} and \cite{pei2020geom}. There are ten different splits for train, validation, and testing. Data profiles including references can be found in Appendix \ref{appendix_data}. 

\subsection{Expressive power}

\begin{figure}[t]
    \centering
    \includegraphics[width=.5\textwidth]{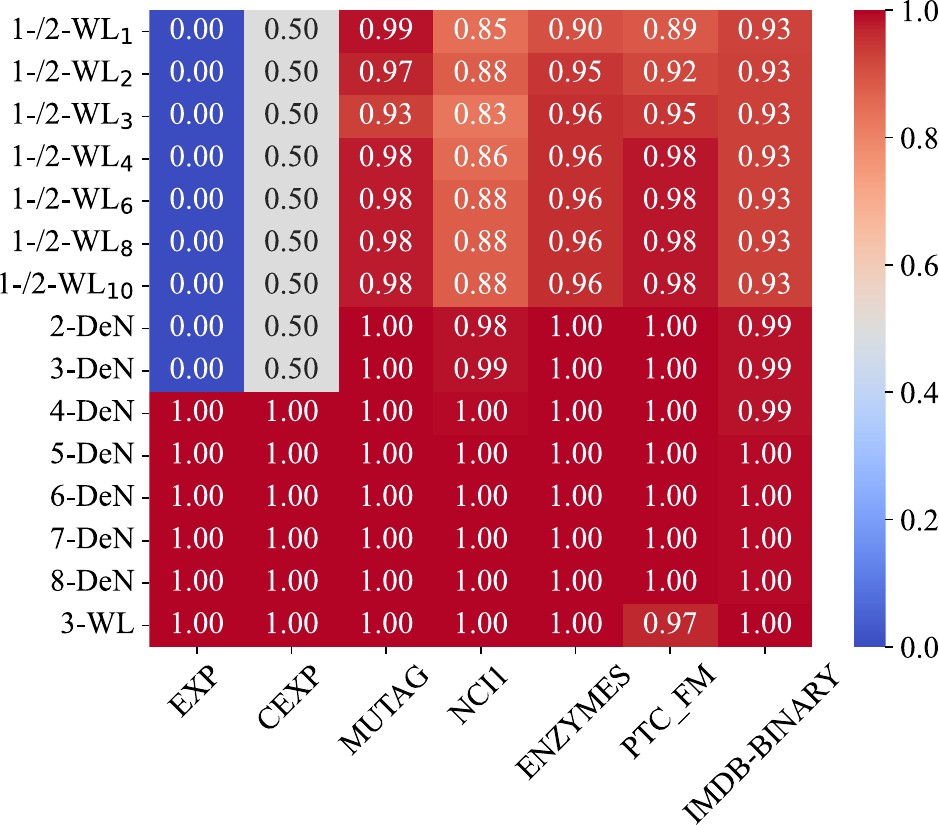}
    \vspace{-2.5em}
    \caption{Expressive power: the ratio of distinguishable graph that is non-isomorphic from pre-defined opponent of synthetic graphs (first two columns from left to right) and graphs have a different class of real-world graphs (last five columns). Each row is sorted based on computational cost. The last row is labeling nodes by 1-WL results of every 3-tuple and other rows are all labeling by a node-wise representation. The subscript of row names denotes the number of iterations, and the default is ten.}
    \label{fig:exp-express}
    \vspace{-1.5em}
\end{figure}

\begin{figure*}[t]
    \vspace{-0.5em}
    \centering
    \includegraphics[width=1.17\textwidth]{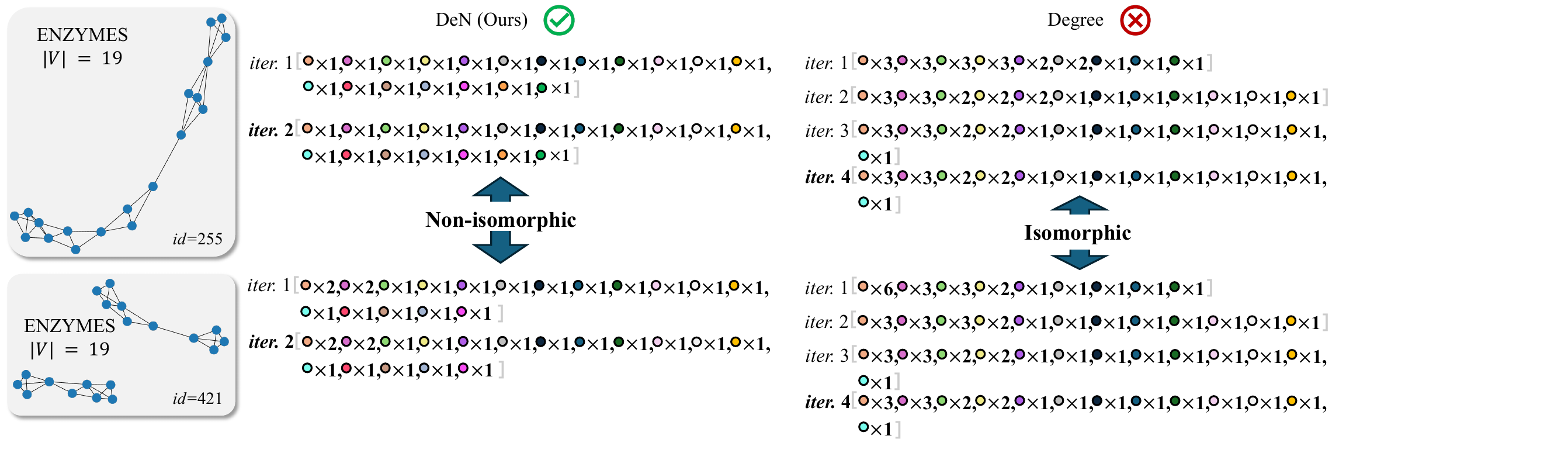}
    \vspace{-3em}
    \caption{Coloring results of each iteration of 1-WL tests for one example of a pair of real-world graphs from the ENZYMES dataset. Degree-initialized WL failed to distinguish this pair, while the proposed detour number (DeN) succeeded.}
    \label{fig:wl_fail_exp}
    \vspace{-1em}
\end{figure*}

In practice, the expressive power is denoted by the ratio of distinguishable graphs that are non-isomorphic from the opponent, which has been pre-defined in synthetic datasets. For real-world data, a graph should be recognized to be non-isomorphic with graphs having a different class. Following this notion, we pair each graph with all others belonging to different classes. If there is a computational method can successfully yield non-isomorphic results for all possible pairings, we tag the underlying graph being distinguishable.

The expressive power comparison is shown in Fig. \ref{fig:exp-express}, where $k$-DeN (in Theorem \ref{detouring_pw}) stands for our method. Subscripts of row names are the maximum number of iterations, other rows without a subscript have the maximum of iterations as ten. For synthetic datasets, every graph of EXP and CEXP has cycles with a length longer than 5. According to their profiles of cycle components (see Appendix \ref{appendix_data}), $k$-DeN is able to express the cyclic difference when $k\geq 4$, where it achieves 100\% distinguishable graph pairs of EXP and CEXP listed in Fig. \ref{fig:exp-express}.

Graphs of real-world datasets are firstly filtered to remove isomorphic graphs and then paired across classes. Graphs that have different classes should be discriminated against each other via topological information, and consequentially, a million-level number of pairs are produced in these datasets. Unlike synthetic data being strictly indistinguishable by 1-/2-WL, 1- or 2-WL test works for the majority of real-world graphs. As shown in Fig. \ref{fig:exp-express}, our method can 100\% distinguish such graphs if $k\geq 5$, while at most 12\% of real-world graphs cannot be passed by the original WL tests. As the conclusion, the answer of \textbf{Q1} is yes.


\begin{table*}[h]
    \centering
    \caption{Graph classification accuracy. Neural networks all have 4 layers to show a fair comparison. The first place is bold and the second is underlined. Accuracy is colored green if it is improved by our proposal (bold methods). `*' means the improvement is higher than 4\%.}
    \small
    \begin{sc}
    \begin{tabular}{l|ccccccc}\toprule
                    & MUTAG & NCI1 & ENZYMES & DD & PTC\_FM & IMDB-B \\\hline\hline
        Graph Kernel \\\hline
        Shortest Path & 77.62$_{\pm1.60}$ & 67.09$_{\pm0.21}$ & 27.60$_{\pm1.12}$ & 74.01$_{\pm0.69}$ & 59.38$_{\pm1.73}$ & 59.09$_{\pm0.64}$\\
        WL Subtree & 79.05$_{\pm1.62}$ & 81.58$_{\pm0.13}$ & 39.87$_{\pm0.89}$ & 74.38$_{\pm0.26}$ & 58.41$_{\pm1.09}$ & 73.67$_{\pm0.20}$\\
        WL SP & 81.09$_{\pm1.62}$ & \textbf{82.81}$_{\pm0.08}$ & 40.52$_{\pm1.28}$ & OOM & 59.50$_{\pm1.21}$ & \underline{74.86}$_{\pm0.28}$\\
        \textbf{WL DeN} (\textcolor{green}{$\uparrow$ 1.06}) & \cellcolor{green!25}\underline{85.10}$_{\pm 2.28}$ & \underline{82.03}$_{\pm0.18}$ & \cellcolor{green!25}41.35$_{\pm1.14}$ & \cellcolor{green!25}\textbf{75.50}$_{\pm 0.51}$ & \cellcolor{green!25}59.89$_{\pm 1.16}$ & \cellcolor{green!25}\textbf{75.63}$_{\pm 0.31}$\\\hline\hline
        MPNN \\\hline
         GCN & 71.54$_{\pm8.85}$ & 77.86$_{\pm2.42}$ & 51.83$_{\pm9.08}$ & 69.52$_{\pm 4.63}$ & 60.98$_{\pm 7.23}$ & 70.20$_{\pm 2.71}$\\
         GAT & 77.18$_{\pm10.63}$ & 68.86$_{\pm3.06}$ & 49.50$_{\pm7.96}$ & \underline{73.00}$_{\pm4.52}$ & \underline{63.11}$_{\pm 7.84}$ & 71.30$_{\pm2.87}$\\
         GIN & 81.79$_{\pm 12.21}$ & 80.82$_{\pm 2.24}$ & 59.66$_{\pm 6.70}$ & 70.28$_{\pm 4.05}$ & 62.16$_{\pm 6.64}$ & 69.00$_{\pm 3.61}$\\\hline\hline
        \textbf{MDNN} \\\hline
         GCN (\textcolor{green}{$\uparrow$ 1.68}) & \cellcolor{green!25}74.40$_{\pm 12.17}$ & 76.11$_{\pm 1.05}$ & \cellcolor{green!25}\underline{59.33}$^*_{\pm 7.20}$ & \cellcolor{green!25}71.21$_{\pm 4.09}$ & 60.63$_{\pm 5.79}$ & \cellcolor{green!25}70.30$_{\pm 3.20}$\\
         GAT (\textcolor{green}{$\uparrow$ 2.39}) & \cellcolor{green!25}81.41$^*_{\pm 10.25}$ & \cellcolor{green!25}77.03$^*_{\pm 1.36}$ & \cellcolor{green!25}55.50$^*_{\pm 4.95}$ & 68.92$_{\pm 3.12}$ & 61.93$_{\pm 7.27}$ & \cellcolor{green!25}72.50$_{\pm 3.23}$\\
         GIN (\textcolor{green}{$\uparrow$ 1.99}) & \cellcolor{green!25}\textbf{86.24}$^*_{\pm 9.01}$ & \cellcolor{green!25}81.17$_{\pm 2.30}$ &  \cellcolor{green!25}\textbf{64.33}$^*_{\pm 7.16}$ & \cellcolor{green!25}70.55$_{\pm 5.22}$ & \cellcolor{green!25}\textbf{63.93}$_{\pm 6.94}$ & \cellcolor{green!25}69.40$_{\pm 3.64}$
         \\\bottomrule
    \end{tabular}
    \end{sc}
    \label{tab:graph_cls}
    \vspace{-1.5em}
\end{table*}

\begin{table*}[h]
    \small
    \vspace{-1em}
    \centering
    \caption{Node classification accuracy. Neural networks all have 4 layers to show a fair comparison. The first place is bold and the second is underlined. Accuracy is colored green if it is improved by our proposal (bold methods). Star means the improvement is higher than 4\%.}
    \begin{sc}
    \begin{tabular}{l|cccccc}\toprule
             & Chameleon & Squirrel & Actor & Cornell & Wisconsin & Texas \\\hline\hline
        MPNN \\\hline
         GCN & 33.62$_{\pm 2.28}$ & 26.05$_{\pm 1.42}$ & 27.43$_{\pm 0.99}$ & 39.19$_{\pm 6.54}$ & 48.63$_{\pm 5.53}$ & 57.30$_{\pm 3.59}$ \\
         GAT  & 40.11$_{\pm 1.67}$ & 28.20$_{\pm 1.56}$ & 27.87$_{\pm 0.95}$ & 41.62$_{\pm 6.86}$ & 51.18$_{\pm 4.59}$ & 55.68$_{\pm 4.22}$ \\
         GIN  & 40.95$_{\pm 2.48}$ & 29.10$_{\pm 2.59}$ & 26.00$_{\pm 1.36}$ & 30.27$_{\pm 7.13}$ & 35.29$_{\pm 6.62}$ & 40.81$_{\pm 14.53}$\\\hline\hline
        \textbf{DeN-weighted} \\\hline
         GCN (\textcolor{green}{$\uparrow$ 2.06}) & 32.74$_{\pm 2.24}$ & \cellcolor{green!25}31.18$^*_{\pm 1.97}$ & \cellcolor{green!25}28.78$_{\pm 1.02}$ & \cellcolor{green!25}41.35$_{\pm 4.84}$ & \cellcolor{green!25}{52.16}$_{\pm 5.90}$ & \cellcolor{green!25}\underline{58.38}$_{\pm 5.82}$\\
         GAT (\textcolor{green}{$\uparrow$ 4.20}) & \cellcolor{green!25}41.16$_{\pm 1.88}$ & \cellcolor{green!25}40.19$^*_{\pm 1.34}$ & \cellcolor{green!25}28.48$_{\pm 0.80}$ & \cellcolor{green!25}45.41$_{\pm 6.60}$ & \cellcolor{green!25}{57.06}$^*_{\pm 3.87}$ & \cellcolor{green!25}57.57$_{\pm 6.95}$\\
         GIN (\textcolor{green}{$\uparrow$ 0.12}) & 27.24$_{\pm 1.95}$ & 24.53$_{\pm 2.71}$ & 25.16$_{\pm 2.17}$ & \cellcolor{green!25}38.92$^*_{\pm 7.76}$ & \cellcolor{green!25}42.94$^*_{\pm 5.71}$ & \cellcolor{green!25}44.32$_{\pm 10.05}$\\\hline\hline
        \textbf{MDNN} \\\hline
         GCN (\textcolor{green}{$\uparrow$ 9.61}) & \cellcolor{green!25}49.19$^*_{\pm 1.47}$ & \cellcolor{green!25}36.95$^*_{\pm 1.32}$ & \cellcolor{green!25}\textbf{32.72}$^*_{\pm 1.18}$ & \cellcolor{green!25}\textbf{52.70}$^*_{\pm 7.18}$ & \cellcolor{green!25}\underline{58.04}$^*_{\pm 11.20}$ & \cellcolor{green!25}\textbf{60.27}$_{\pm 7.46}$ \\
         GAT (\textcolor{green}{$\uparrow$ 6.73}) & \cellcolor{green!25}\underline{52.59}$^*_{\pm 2.53}$ & \cellcolor{green!25}\underline{38.68}$^*_{\pm 0.95}$ & \cellcolor{green!25}31.82$_{\pm 1.44}$& \cellcolor{green!25}50.00$^*_{\pm 6.86}$ & \cellcolor{green!25}56.08$^*_{\pm 8.95}$ & \cellcolor{green!25}56.76$_{\pm 9.28}$ \\
         GIN (\textcolor{green}{$\uparrow$ 12.86}) & \cellcolor{green!25}\textbf{54.52}$^*_{\pm 2.06}$ & \cellcolor{green!25}\textbf{39.72}$^*_{\pm 1.54}$ & \cellcolor{green!25}\underline{32.28}$^*_{\pm 0.72}$ & \cellcolor{green!25}\underline{51.89}$^*_{\pm 10.32}$ & \cellcolor{green!25}\textbf{67.65}$^*_{\pm 7.50}$ & \cellcolor{green!25}56.76$^*_{\pm 6.84}$
         \\\bottomrule
    \end{tabular}
    \end{sc}
    \label{tab:node_cls}
    \vspace{-1.5em}
\end{table*}

Take one pair of graphs of ENZYMES as an example shown in Fig. \ref{fig:wl_fail_exp}. In the left part, the number of components is an obvious high-dimensional element to recognize the isomorphism, while node-wise representation by 1-WL cannot obtain such information as shown in the Fig. \ref{fig:wl_fail_exp} right. However, the meaningful feature is the cycle counts. The final coloring results as shown in the Fig. \ref{fig:wl_fail_exp} middle inherit the difference in the cycle pattern. The upper graph has two more colors than the lower graph by DeN. That highlights the advantage of our proposal as a node-wise cycle representation to recognize such graphs driven by high-dimensional information.

\subsection{Accuracy of graph classification}
The more expressive power of DeN should lead to a better performance. Firstly, we test our three implementations of message detouring on the graph classification task with six real-world datasets. As listed in Table \ref{tab:graph_cls}, implementations are deployed under four baselines, WL shortest path (WL SP) kernel, graph convolution network (GCN), graph attention network (GAT), and graph isomorphism network (GIN). These four baselines are common kernel and nueral approaches for graph embedding, where graph kernel and GIN work better than GCN and GAT on graph classification \cite{siglidis2020grakel,errica2019fair}. 
Due to space limits, we put results of DeN-weighted MPNNs as supplementary in Appendix \ref{appendix_supp_res}.

The graph kernel part of Table \ref{tab:graph_cls} compares three different kernels, Shortest Path (SP), WL Subtree, and WL SP following \cite{shervashidze2011weisfeiler,siglidis2020grakel}. Ours WL DeN initializes the best baseline, WL SP, using the proposed $\{\!\!\{D, k$-DeN\}\!\!\} with $k$ set properly according to the cycle length of the data. It is evident that five out of six datasets have improved by using the node-wise cycle representation. Accuracy is increased by 1.06\% on average. 

The MPNN part of Table \ref{tab:graph_cls} benchmarks neural approaches under our experimental environments. Ranks of neural baselines are consistent with related literature that GIN is the best for this task, while four of the datasets are lower than our graph kernel method, e.g., WL DeN is 3.31\% more accurate than GIN for MUTAG.


The MDNN panel lists the results of the proposed Transformer-style message detouring. With the exception of four out of eighteen results, the remaining outcomes show improvements, with half of them exceeding a 4\% improvement compared to the baselines. For example, GAT trained with MDNN feature, that is GAT under MDNN, on NCI1 is 8.17\% higher than MPNN.

To conclude experiments of graph classification and answer \textbf{Q2}, the proposed detour number and message detouring show significant effectiveness. The performance of baselines can be enhanced by our node-wise cycle representation.

\subsection{Accuracy of node classification}
Utilizing the structural representation, for example, cycle in this work, for node-level tasks is commonly by graph similarity, such as struc2vec \cite{ribeiro2017struc2vec}, but it is rare by neural networks. 
We compare neural network counterparts with our MDNN-bssed deep model on six graph datasets where the rich structural information (see their cycle profiles in Appendix \ref{profile}) may benefit graph learning. We exclude graph kernels since they are not capable of node classification.

As listed in Table \ref{tab:node_cls}, the same GNN baselines are deployed. To summarize and answer \textbf{Q3}, both DeN-weighted and MDNN implementations of our proposal show significant improvement in node classification accuracy, where our MDNN shows the most promising results comparing baselines. For example on the small-scale graph Wisconsin, GIN gains 37.38\% more accuracy by using new feature by MDNN. Besides, 13.57\% additional accuracy is attained by GIN+MDNN in the big-scale graph Chameleon.

The contribution associated with DeN-weighted part (using $k$-DeN as the edge weight) leads to improved node classification performance in fourteen out of eighteen datasets. In contrast to our experiments on the graph-level task, we find that GNI+DeN-weighted MPNN has achieved significant improvement on small-scale datasets such as Cornell, Wisconsin, and Texas while GIN alone does not work well on these datasets. 
Similarly, DeN-weighted GCN shows the second best accuracy 58.38\% on Texas data.

MDNN on node classification achieves significant performance enhancement to the extent that (1) half of green results are higher than 4\% and (2) the average of improvement reaches 12.86\% for GIN using the new feature from MDNN. This provides another piece of strong evidence of the positive effect of node-wise cycle representation for node classification.

Referring to the cycle profiles of various datasets, message detouring can unleash more performance on node classification no matter the graph has massive cycles like Actor contains more than 25K cycles with length no longer than 5 and Chameleon has millions of such cycles, or graphs with few cycles like Texas less than 7 cycles only.

\subsection{Ablation studies}
Ablation studies are conducted to show the detour embedding $\varphi_D$ of MDNN is effective. As listed in Table \ref{tab:ablation}, where MUTAG is used for graph classification, Chameleon and Cornell and Texas are used for node classification. It is clear the detour embedding has substantial contribution that leads to higher performance.

\begin{table}[h]
    \centering
    \small
    \vspace{-1.5em}
    \caption{Ablation studies for MDNN with the same setting as main experiments. The baseline here is a 4-layer GAT.}
    \begin{sc}
\setlength{\tabcolsep}{2pt} 
    \begin{tabular}{p{0.07\textwidth}|cccc}\toprule
    & MUTAG & Chameleon & Cornell & Texas \\\hline
    w/o $\varphi_D$ & 79.57$_{\pm 7.95}$ & 50.22$_{\pm 1.99}$ & 45.68$_{\pm 7.95}$  & 54.05$_{\pm 9.52}$\\
    w/ $\varphi_D$ & 81.41$_{\pm 10.25}$ & 52.59$_{\pm 2.53}$ & 50.00$_{\pm 6.86}$  & 56.76$_{\pm 9.28}$\\\bottomrule
    \end{tabular}
    \end{sc}
    \label{tab:ablation}
    \vspace{-1em}
\end{table}

\section{Related works}



Graph topology methods, such as CBGNN \cite{yan2022cycle}, CycleNet \cite{yan2023cycle}, GSN \cite{bouritsas2022improving}, and many other works \cite{arvind2020weisfeiler,chen2020can} study a variety of substructures for graph to graph comparison, not limited to cycle only. However, very few attentions have been paid to characterize the topological characteristics at the node level. In terms of practical application, our MDNN method has greater scalability, effectively encompassing graph learning applications at both the node and graph levels.

Tuple-wise representation works usually use high-order models such as path representation \cite{fu2020magnn,chen2021combined,michel2023path}, hypergraph \cite{chen2020can} \cite{zhao2022practical}, or subgraph GNNs \cite{zhao2021stars,bevilacqua2021equivariant,frasca2022understanding,qian2022ordered}. Path representation technique has been extended to meta-path \cite{fu2020magnn,chen2021combined}, and shortest path \cite{michel2023path}. In contrast to our MDNN method, these high-order models often require significantly greater computational resources. 

Many recent works on node-wise representation learning have been proposed, including node-to-node dynamics \cite{wang2022acmp}, remapping geometric space \cite{pei2020geom}, graph structure learning \cite{wu2022nodeformer}, and graph rewiring \cite{guo2023homophily}. Particularly, Relative Random Walk Probabilities (RRWP) \cite{ma2023graph} replaces the node connectivity degree with a random walk, which is very similar to our MDNN where we use detour path. RRWP can be theoretically generalized to any substructure, while the detour path in our explicitly focuses on cycles in the graph.








\section{Conclusion}

In this work, we introduced the concept of message detouring to the field of expressive graph learning. We first propose an edge-wise feature to represent the cycle via the number of detour paths. We provide theoretical support to show that detour number is more expressive, and can be as powerful as high-dimensional WL tests in specific cases with lower computational costs. Three practical implementations are provided to frame message detouring, namely, WL DeN, DeN-weighted MPNN, and a novel message detouring neural network using Transformer. 
Experimental results show strong evidence of the expressive power of DeN by 100\% distinguishing millions of non-isomorphic graph pairs of synthetic and real-world datasets. On the other hand, our results of real-world tasks of graph classification and node classification shows significant improvement in terms of accuracy, compared with message passing counterpart methods. These promising results demonstrate the advantage of message detouring in recognizing cycles in graph learning. 


\section{Impact Statement}
This paper presents work whose goal is to advance the field of Machine Learning. There are many potential societal consequences of our work, none which we feel must be specifically highlighted here.



\bibliography{example_paper}
\bibliographystyle{icml2024}

\newpage
\appendix
\onecolumn

\section{Accessibility}

All data and split settings can be accessible via the Pytorch-Geometric library\footnote{\url{https://pytorch-geometric.readthedocs.io/en/latest/}}, TUDataset\footnote{\url{https://chrsmrrs.github.io/datasets/docs/datasets/}}, and this GitHub repository\footnote{\url{https://github.com/diningphil/gnn-comparison/tree/master/data_splits}}. 

\section{Data profiles}\label{appendix_data}
\label{profile}
Data profiles of 12 real-world datasets are listed in this section. Basic profiles can be found in Table \ref{tab:data_prof}, where \textit{homo. ratio} stands for the ratio of homophilous graphs \cite{zhu2020beyond}. 

\begin{table}[h]
    \centering
    \small
    \vspace{-2em}
    \caption{Data profiles of real-world datasets.}
    \begin{sc}
    \begin{tabular}{l|p{0.12\linewidth}p{0.12\linewidth}p{0.12\linewidth}p{0.12\linewidth}p{0.12\linewidth}p{0.12\linewidth}p{0.12\linewidth}}\toprule
    & MUTAG & NCI1 & ENZYMES & DD & PTC\_FM & IMDB-B \\
    & \cite{kriege2012subgraph} & \cite{shervashidze2011weisfeiler} & \cite{schomburg2004brenda}  & \cite{shervashidze2011weisfeiler} & \cite{kriege2012subgraph} & \cite{yanardag2015deep} \\\hline
    $avg(|V|)$ & 17.93 & 29.87 & 32.63 & 284.32 & 14.11 & 19.77\\
    $avg(|E|)$ & 19.79 & 32.30 & 62.14 & 715.66 & 14.48 & 96.53\\
    $|G|$ & 188 & 4110 & 600 & 1178 & 349 & 1000\\
    \textit{class num.} & 2 & 2 & 6 & 2 & 2 & 2 \\\hline\hline
    & Chameleon & Squirrel & Actor & Cornell & Wisconsin & Texas \\
    & \cite{rozemberczki2021multi} & \cite{rozemberczki2021multi} & \cite{pei2020geom} & \cite{pei2020geom}& \cite{pei2020geom}& \cite{pei2020geom}\\\hline
    $|V|$ & 2277 & 5201 & 7600 & 183 & 251 & 183 \\
    $|E|$ & 36101 & 217073 & 30019 & 298 & 515 & 325 \\
    $avg(D)$ & 15.85 & 41.74 & 3.95 & 1.63 & 2.05 & 1.78 \\
    \textit{class num.} & 5 & 4 & 5 & 5 & 5 & 5 \\
    \textit{homo. ratio} & 0.235 & 0.224 & 0.219 & 0.131 & 0.196 & 0.108\\\bottomrule
    \end{tabular}
    \end{sc}
    \label{tab:data_prof}
\end{table}

\subsection{Cycle components of graphs}

Graphs in our experiments have various cases of cycles. We skipped cycles have longer than 6 paths (or 5 for Squirrel and Chameleon) for each dataset as listed in Table \ref{tab:cyc_dist}. For graphs in TUDataset, we skipped graphs that have a $k$-factor spanning subgraph if $k>2$ to avoid counting cycles on regular subgraphs.

\begin{table}[h]
    \centering
    \small
    \caption{Data profiles of cycle components. The cycle length combination is denoted by a single color for each dataset, e.g., EXP has 888 graphs containing multiple cycles, where each graph has at least three cycle length equal to 4, 5, and 6, respectively. Cells without coloring is the number of cycles for single-graph datasets. '-' means is not applicable.}
    \begin{sc}
    \begin{tabular}{l|l|p{0.045\linewidth}|p{0.045\linewidth}|p{0.045\linewidth}|p{0.045\linewidth}||l|l|p{0.045\linewidth}|p{0.045\linewidth}|p{0.045\linewidth}|p{0.045\linewidth}}\toprule
     & Non- & \multicolumn{4}{c||}{\textit{cycle length}=} & & Non- & \multicolumn{4}{c}{\textit{cycle length}=} \\\cline{3-6}\cline{9-12}
        Data name & cyclic & 3 & 4 & 5 & 6 & Data name & cyclic & 3 & 4 & 5 & 6 \\\hline
        EXP & 0 & & \multicolumn{1}{l}{\cellcolor{gray!25}888} & \multicolumn{2}{l||}{\cellcolor{gray!25}} & PTC\_FM & 42 & \multicolumn{1}{l|}{\cellcolor{blue!10}3} & \cellcolor{red!30}1
 & \multicolumn{2}{l}{\cellcolor{blue!10}}\\
         && & \cellcolor{red!25}26 & & \cellcolor{red!25}  &  & & \cellcolor{green!30}2 & \multicolumn{3}{l}{\cellcolor{gray!30}3}\\
         && & & \multicolumn{1}{l}{\cellcolor{blue!25}284} & \cellcolor{blue!25}  &&& \cellcolor{purple!25}2 & \cellcolor{gray!50}1 & \cellcolor{yellow!50}8& \cellcolor{purple!25}\\
         && & & & \cellcolor{green!25}4  &&&& &\multicolumn{2}{l}{\cellcolor{blue!50}28}\\
         && & & &   &&& & & & \cellcolor{purple!50}153\\
         \hline
        CEXP  &0& & \multicolumn{1}{l}{\cellcolor{gray!25}932} & \multicolumn{2}{l||}{\cellcolor{gray!25}} & DD & 5 & \multicolumn{4}{l}{\cellcolor{gray!25}1173}\\
         && & \cellcolor{red!25}8 & & \cellcolor{red!25}  & &&&&&\\
         && & & \multicolumn{1}{l}{\cellcolor{blue!25}255} & \cellcolor{blue!25}  &&&&&&\\
         && & & & \cellcolor{green!25}5  &&&&&&\\
         \hline
        MUTAG  &0& & \cellcolor{red!25}2 & \multicolumn{1}{l}{\cellcolor{gray!25}44} & \multicolumn{1}{l||}{\cellcolor{gray!25}} & IMDB-B & 194 & \multicolumn{4}{l}{\cellcolor{gray!25}297}\\
         & & & & \cellcolor{blue!25}89 & & & & \multicolumn{3}{l}{\cellcolor{blue!25}2} &\\
         \hline
        NCI1  & 103 & \multicolumn{4}{l||}{\cellcolor{gray!25}1} & ENZYMES & 80 & \multicolumn{4}{l}{\cellcolor{gray!25}504}\\
         & & & \multicolumn{3}{l||}{\cellcolor{red!25}23} & & & \multicolumn{3}{l|}{\cellcolor{blue!25}7} &\\
         & & & & \multicolumn{2}{l||}{\cellcolor{blue!25}1586} & & & \cellcolor{red!25}1 &\multicolumn{3}{l}{\cellcolor{yellow!25}1}\\
         & & \multicolumn{1}{l|}{\cellcolor{blue!10}78} & & \multicolumn{2}{l||}{\cellcolor{blue!10}} & &&\multicolumn{2}{l|}{\cellcolor{purple!25}1} & & \cellcolor{purple!25}\\
         & & \multicolumn{1}{l|}{\cellcolor{yellow!90}8} & \multicolumn{1}{l|}{\cellcolor{red!10}67} & \multicolumn{1}{l|}{\cellcolor{yellow!90}} & \multicolumn{1}{l||}{\cellcolor{red!10}} & &&\multicolumn{2}{l|}{\cellcolor{gray!50}1} & & \\
         & & \multicolumn{1}{l|}{\cellcolor{purple!50}41} & \multicolumn{2}{l|}{\cellcolor{green!30}3} & \multicolumn{1}{l||}{\cellcolor{purple!50}} & &&&&&\\
         & & \cellcolor{gray!40}7 & \cellcolor{blue!40}7 & \cellcolor{red!40}141 & \cellcolor{yellow!40}1720 & &&&&&\\
         \hline
        Squirrel & - & \multicolumn{1}{@{}l|}{\ 1.2M} & \multicolumn{1}{@{}l|}{\ 111.3M} & \multicolumn{1}{@{}l|}{\ 11.3B} & - & Cornell & - & 6 & 4 & 2 &0\\
         \hline
        Chameleon & - & \multicolumn{1}{@{}l|}{\ 124.7K} & \multicolumn{1}{@{}l|}{\ 5.0M} & \multicolumn{1}{@{}l|}{\ 220.8M} & - & Wisconsin &-& 10 & 11 & 2 &0\\
         \hline
        Actor & - & \multicolumn{1}{@{}l|}{\ 1.5K} & \multicolumn{1}{@{}l|}{\ 4.0K} & \multicolumn{1}{@{}l|}{\ 20.8K} &0&Texas&-& 6 & 1 & 0& 0\\
         \bottomrule
    \end{tabular}
    \end{sc}
    \label{tab:cyc_dist}
\end{table}

\section{Detour paths in Fig. \ref{fig:wl_failure}}
\label{appendix_detour_path}

\begin{figure}[h]
    \includegraphics[width=0.7\textwidth,center]{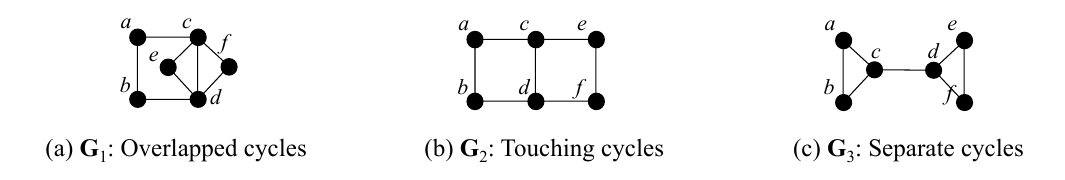}
    \vspace{-1em}
    \caption{Recap of Fig. \ref{fig:wl_failure}.}
    \vspace{-5em}
\end{figure}

\subsection{Detour paths in graph $\mathbf G_1$}
\begin{enumerate}
    \item Edge $ab$: \#1 $\overrightarrow{acdb}$
    \item Edge $ac$: \#1 $\overrightarrow{abdc}$
    \item Edge $bd$: \#1 $\overrightarrow{bacd}$
    \item Edge $cd$: \#1 $\overrightarrow{cabd}$, \#2 $\overrightarrow{ced}$, \#3 $\overrightarrow{cfd}$
    \item Edge $ce$: \#1 $\overrightarrow{cde}$, \#2 $\overrightarrow{cfde}$
    \item Edge $cf$: \#1 $\overrightarrow{cdf}$, \#2 $\overrightarrow{cedf}$
    \item Edge $de$: \#1 $\overrightarrow{dce}$, \#2 $\overrightarrow{dfce}$
    \item Edge $df$: \#1 $\overrightarrow{dcf}$, \#2 $\overrightarrow{decf}$
\end{enumerate}

\subsection{Detour paths in graph $\mathbf G_2$}

\begin{enumerate}
    \item Edge $ab$: \#1 $\overrightarrow{acdb}$
    \item Edge $ac$: \#1 $\overrightarrow{abdc}$
    \item Edge $bd$: \#1 $\overrightarrow{bacd}$
    \item Edge $cd$: \#1 $\overrightarrow{cabd}$, \#2 $\overrightarrow{cefd}$
    \item Edge $ce$: \#1 $\overrightarrow{cdfe}$
    \item Edge $df$: \#1 $\overrightarrow{dcef}$
    \item Edge $ef$: \#1 $\overrightarrow{ecdf}$
\end{enumerate}

\subsection{Detour paths in graph $\mathbf G_3$}

\begin{enumerate}
    \item Edge $ab$: \#1 $\overrightarrow{acb}$
    \item Edge $ac$: \#1 $\overrightarrow{abc}$
    \item Edge $bc$: \#1 $\overrightarrow{bac}$
    \item Edge $cd$: None
    \item Edge $de$: \#1 $\overrightarrow{dfe}$
    \item Edge $df$: \#1 $\overrightarrow{def}$
    \item Edge $ef$: \#1 $\overrightarrow{edf}$
\end{enumerate}

\section{Proof}\label{proof_theorem}

\subsection{Proposition \ref{den_cycle_counting}}
\label{proof_den_cycle_counting}

\textbf{Proposition \ref{den_cycle_counting}.} 
\textit{$\frac{1}{2}\Phi^{k}_i$ characterizes the cardinality of the set of cycles with length no longer than $(k+1)$ where each cycle includes node $V_i$.} 

\begin{proof}
A cycle $\Lambda$ can be separate as a detour path $\mathcal P_{ab}$ and the edge $E_{ab}$ given a node $V_a$ and $b\in\mathcal N(V_a)$ 
\begin{equation}\label{eq_lambda}
    \Lambda_{ab}:=\mathcal P_{ab}\cup{E_{ab}}
\end{equation}
then, based on the Definition \ref{def1}, the set of cycles with length no longer than $(k+1)$ can be denoted by $$\mathbf{C}(V_a, k+1):=\{\Lambda_{ab}|\forall b\in\mathcal N(V_a),\exists\mathcal P_{ab},2<|\Lambda_{ab}| \leq k+1\}$$ with respect to node $V_a$. Obviously, $1<|\mathcal P_{ab}|\leq k$, and $|\mathbf{C}(V_a, k+1)| = |\{\Lambda_{ab}\}|$ after removing duplicated $\Lambda$. Since it is easy to see that at least three nodes $V_a, V_b, V_c$ exist to construct a cycle where at least two edges exist to connect the three so that two detour paths exist in one cycle:
$$\forall \mathcal P_{ab},\ \exists c\in\mathcal P_{ab}, c\in\mathcal N(V_a)\Rightarrow$$
$$\exists\mathcal P_{ac},b\in\mathcal P_{ac}.$$
Then, we can have
$$\mathcal P_{ab}\setminus E_{ac}=\mathcal P_{ac}\setminus E_{ab}=\mathcal P_{bc}\Rightarrow$$
\begin{equation}\label{eq_path}
    \begin{cases}
    \mathcal P_{ab}=\mathcal P_{bc}\cup \{E_{ac}\}\\
    \mathcal P_{ac}=\mathcal P_{bc}\cup \{E_{ab}\}
    \end{cases}.
\end{equation}
Thus, combine Equation \ref{eq_lambda} and Equation \ref{eq_path}, we get
\begin{equation*}
    \begin{cases}
        \Lambda_{ab} := \mathcal P_{ab}\cup\{E_{ab}\} = \mathcal P_{bc}\cup \{E_{ac}\} \cup\{E_{ab}\}\\
        \Lambda_{ac} := \mathcal P_{ac}\cup\{E_{ac}\} = \mathcal P_{bc}\cup \{E_{ab}\} \cup\{E_{ac}\}\\
    \end{cases}\Rightarrow
\end{equation*}
\begin{equation}\label{eq_lambda_equiv}
    \begin{cases}
      \Lambda_{ab}\equiv\Lambda_{ac}\\
      \mathcal P_{ab}\not\equiv\mathcal P_{ac}
    \end{cases}\,.
\end{equation}

Based on Definition \ref{def1} and \ref{def2}, $k\text{-}DeN=|\{\mathcal P|1<|\mathcal P|\leq k\}|$ in a short notation, and in cases of Equation \ref{eq_lambda_equiv} we know $|\{\Lambda_{ab},\Lambda_{ac}\}|=1$ and $|\{\mathcal P_{ab},\mathcal P_{ac}\}|=2|\{\Lambda_{ab},\Lambda_{ac}\}|=2$ since $V_a, V_b, V_c$ always exist in a cycle. To this end, in brief, we get $\sum k\text{-}DeN=|\{\mathcal P\}|=2|\{\Lambda\}|=2|\mathbf{C}|$, that is $$2|\mathbf{C}(V_a, k+1)|=\sum_{b\in\mathcal N(V_a)}k\text{-}DeN_{ab}=\Phi^{k}_a$$ This finishes the proof.
\end{proof}

\subsection{Theorem \ref{detouring_pw}}
\label{proof_pw}

\textbf{Theorem \ref{detouring_pw}.} 
\textit{$\{\!\!\{D(V_i)\}\!\!\}\cup\{\!\!\{k\text{-}DeN_{ij}|j\in\mathcal N(V_i)\}\!\!\}$ (be shorten as $\{\!\!\{D,k\text{-}DeN\}\!\!\}$ in the later text)-initialized $1$-WL, if used to express isomorphism of connected graphs, is stronger than the degree-initialized $1$-WL, and can reach the expressive power of $k$-WL.}

\subsubsection{Notations}
Given a connected graph $G=\{V, E\}$ with cycle set denoted by $\mathbf C$, where $k+1$ the length of the longest cycle, and $\overline G=\{\overline V, \overline E\}$ is one connected non-isochronic graph with cycle set $\overline{\mathbf C}$. Degree is the scalar of star-shaped structure for a node, i.e., $S^*(v_i):=D(v_i)$. Detour number is the scalar of cyclic structure for a node, i.e., $S^c(v_i):=\{\!\!\{D(V_i)\}\!\!\}\cup\{\!\!\{k\text{-}DeN_{ij}|j\in\mathcal N(V_i)\}\!\!\}$. We denote $\text{hash}(S^c)\equiv\text{hash}(S^*)$ by $S^c\Leftrightarrow S^*$.
 
 Before proving the Theorem \ref{detouring_pw}, we first propose a Lemma states the condition of equivalent expressive power between the degree and DeN.
\begin{lemma}\label{appendix_lemma}
    If $\exists t_1\leq T_1, t_2\leq T_2$, and $\forall v\in V, S^c_{t_1}(v) \Leftrightarrow S^*_{t_2}(v)$, where $T_1,T_2$ are the iteration numbers of convergence, then $T_1-t_1=T_2-t_2:=\delta$, and $\forall i\leq\delta, v\in V, S^c_{t_1+i}(v) \Leftrightarrow S^*_{t_2+i}(v)$. 
\end{lemma}

\begin{wrapfigure}{r}{0.15\textwidth}
\vspace{-1em}
  \begin{center}
    \includegraphics[width=0.135\textwidth]{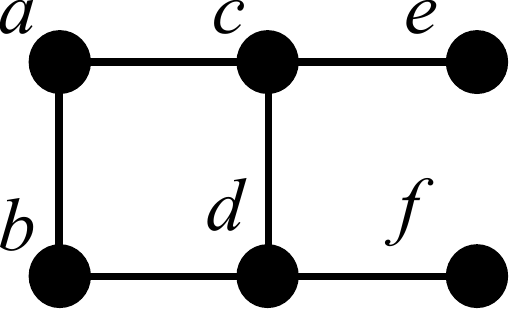}
  \end{center}
\vspace{-1.5em}
  \caption{Tailed cyclic graph.}\label{tail_g}
\end{wrapfigure}

This Lemma leads to the converged coloring results becoming the same between different initializations of $S^*$ and $S^c$, which means the same expressive power. One example is the tailed graph shown in Fig. \ref{tail_g}, where $S^c_0=\{\!\!\{2, 2, 2, 2, 0, 0\}\!\!\}$, $S^*_0=\{\!\!\{2, 2, 3, 3, 1, 1\}\!\!\}$, and $S^c_1=\{\!\!\{222, 222, 2220, 2220, 20, 20\}\!\!\} \Leftrightarrow S^*_0$. Hence, the stronger expressive power derived by opposite cases of Lemma \ref{appendix_lemma}: $\forall t_1\leq T_1, t_2\leq T_2, \exists v\in V, S^c_{t_1}(v) \nLeftrightarrow S^*_{t_2}(v)$ since $S^*$ has no ability of counting cycles.

\begin{proof}
    \textbf{Case 1}: If $\exists v\in V, D(v) = 1$, then $G$ is a tailed graph. For tailed graph, $\exists v_a, v_b, v_c\in V$ are connected, and $k$-DeN$_{ab}=0$ where $D(v_a)\text{ mod }2=1$ and $D(v_b)=1$. In this case, $S^c$ and $S^*$ have the same expressive power for the tailed (sub)graph when $|\mathbf C|=1$. 
    
    The easy case is a cycle without tail, $\forall v_i\in V, D(v_i)=2$ and $k\text{-DeN}_{ij}=1$. As a result, $S^c_{0}(v_i)\Leftrightarrow S^*_{0}(v_i)$. So if tails are appended in this case. Assume $D(v_c)\text{ mod }2=0$, then $S_{0}^c(v_a)=S_{0}^c(v_c)$, but $S_{1}^c(v_c) \neq S_{1}^c(v_a) = \text{HASH}(S_{0}^c(v_a), S_{0}^c(v_b)) \neq S_{0}^c(v_b)=0$. In contrast, $D(v_a)\text{ mod }2=1$, $D(v_b)=1$, and $D(v_c)\text{ mod }2=0$ results $S^*_0(v_a) \neq S^*_0(v_b) \neq S^*_0(v_c)$. Thus,
\begin{equation}
    S^c_{1}\Leftrightarrow S^*_{0},
\end{equation}

\textbf{Case 2}: If $|\mathbf C|\leq1$. In this case other than Case 1, it is easy to see that $\forall i\in V, \sum_{j\in\{\cdot\}^2} k$-DeN$_{ij}=2|\mathbf C|$, since $\forall \{i,j\}\in E, k$-DeN$_{ij}=0$ if $|\mathbf C|=0$ else $k$-DeN$_{ij}=2$. This results $\forall t\leq T, S^c_t\Leftrightarrow S^*_t$.

\textbf{Case 3}: If $|\mathbf C| > 1$, we prove the expressive power by the expression of subgraphs fitting their cycles to Case 1 or Case 2. 

Assume $\mathfrak C$ is the set of cycles $\forall C_1, C_2,  C_3\in\mathbf C$ satisfying $C_1=C_2\cup C_3\setminus C_2\cap C_3$, we get $\forall E_{ab}\in C_2\cap C_3, k$-DeN$_{ab} \geq 3$ and $\forall E_{ab}\in C_1, k$-DeN$_{ab} = 2$. For such $C_1, C_2, C_3$, they can be called touching/overlapped cycles as shown in Fig. \ref{fig:wl_failure}. So that $\sum_{b\in\{\cdot\}^{D(v_a)}} k$-DeN$_{ab} \geq 2D(v_a)$ by adding the detour number for all edges share a various number of cycles. This makes $k$-DeN varies, and hence $S^c_{0}(v) \nLeftrightarrow S^*_{0}(v)$.

For cycles not satisfy the above condition, they as subgraphs, i.e., tailed cycles, are exactly belong to Case 1. According to the same expressive power $S^c_{1}\Leftrightarrow S^*_{0}$ revealed by all subgraphs, the expressive power remains unchanged.

To show the expressive power of $S^c$ at most, we first demonstrate the condition for $z$-WL, i.e., $S^*(V^z)$ where $V^z:=\{v\in V\}^{z}$, to distinguish two non-isomorphic graphs under Case 3 so that $S^{*}(V^z) \nLeftrightarrow S^{*}(\overline{V^z})$. We first show a Lemma in what cases 1-WL can distinguish with a cyclic graph and so that as the input tuple to distinguish two cyclic graphs.
\begin{lemma}\label{appendix_lemma2}
    \textit{If $\forall v\in V,D(v)=2,\exists v\in \overline V,D(v)=1$, then $G$ is a 2-regular graph, i.e., a non-tailed cycle, $\overline G$ is a path or tailed cycle. In this case, $ S^*(V)\nLeftrightarrow S^*(\overline V)$.}
\end{lemma}
Lemma \ref{appendix_lemma2} is easy to prove since $\exists v\in\overline V, D(v)=1$ while the other graph is all 2. Based on this, we can continue to show the expressive power of $S^c(V^z)$ with cyclic graphs. If $C_2$ is the biggest one among interacting cycles, and $\forall V^{|C_2|-1},\overline{V^{|C_2|-1}}, S^{*}(V^{|C_2|-1}) \Leftrightarrow S^{*}(\overline{V^{|C_2|-1}})$. This indicates parts other than cycles within graphs are isomorphic. Then, if and only if $ V^{|C_2|},\overline{V^{|C_2|}}$ exist so that $S^{*}(V^{|C_2|}) \nLeftrightarrow S^{*}(\overline{V^{|C_2|}})$ satisfied since the expressive power of $(z+1)$-WL strictly stronger than $z$-WL \cite{morris2021weisfeiler}. Thus, in the case of $z=|C_2|\equiv (k+1)$, $S^c(V)$ has the same expressive power as $S^{*}(V^z)$. This finishes the proof.
\end{proof}

\section{Implementation details}\label{appendix_imp}

For graph kernels, the prediction is fitted by a Support Vector Machine (SVM) with input from the node labels from kernels. SVMs are fitted using "train+val splits" data and tested on "test split" data for neural approaches to show a fairly comparison.

For all graph neural networks deployed in our experiments, the hidden feature set with 768 channels and 4 layers. Aside from hidden layers, models both have one layer of encoder for input embedding and one layer of decoder for classification. The training loop has the maximum epoch number of 500 with a 250-epoch patience of validation accuracy decreasing. The batch size is set depending on the data size and our hardware. The learning rate is $10^{-5}$ with a warm-up starting. The optimizer is AdamW by PyTorch. 

MDNN on graph classification each multi-head self-attention block has 1 head and $N_{layer}=12$, while on node classification each multi-head self-attention block has 4 heads and $N_{layer}=2$. During training, MDNN fits a batch of graphs as one step on graph classification, while it fits a batch of nodes with their 2-hop neighborhood nodes on node classification. In practice, our model is followed by one of conventional GNN, such as the convolution of GCN \cite{kipf2016semi}, the attentional operator of GAT \cite{velivckovic2017graph}, or the isomorphism operator of GIN \cite{xu2018powerful}. Note that one important difference between the original transformer and MDNN is that the batch normalization (BN) is applied. BN is called since message sequences of the same graph should be normalized together, while natural language processing (NLP) assumes sequences are independent motivating to avoid BN \cite{shen2020powernorm}.

\section{Supplementary results of graph classification}\label{appendix_supp_res}

In this section, we show supplementary results of graph classification in Table \ref{tab:append_sup}.

\begin{table}[h]
    \centering
    \caption{Supplementary results of graph classification. Neural networks all have 4 layers to show a fair comparison. Accuracy is colored green if it is improved by our proposal (bold methods). `*' means the improvement is higher than 4\%.}
    \begin{tabular}{l|ccccccc}\toprule
                    & MUTAG & NCI1 & ENZYMES & DD & PTC\_FM & IMDB-B \\\hline\hline
        MPNN \\\hline
         GCN & 71.54$_{\pm8.85}$ & 77.86$_{\pm2.42}$ & 51.83$_{\pm9.08}$ & 69.52$_{\pm 4.63}$ & 60.98$_{\pm 7.23}$ & 70.20$_{\pm 2.71}$\\
         GAT & 77.18$_{\pm10.63}$ & 68.86$_{\pm3.06}$ & 49.50$_{\pm7.96}$ & \underline{73.00}$_{\pm4.52}$ & \underline{63.11}$_{\pm 7.84}$ & 71.30$_{\pm2.87}$\\
         GIN & 81.79$_{\pm 12.21}$ & 80.82$_{\pm 2.24}$ & 59.66$_{\pm 6.70}$ & 70.28$_{\pm 4.05}$ & 62.16$_{\pm 6.64}$ & 69.00$_{\pm 3.61}$\\\hline\hline
        \textbf{DeN-weighted} \\\hline
         GCN (\textcolor{green}{$\uparrow$ 1.28}) & \cellcolor{green!25}78.08$^*_{\pm9.34}$ & 77.64$_{\pm2.22}$ & \cellcolor{green!25}52.17$_{\pm6.15}$ & \cellcolor{green!25}70.45$_{\pm 4.87}$ & 60.64$_{\pm 8.29}$ & \cellcolor{green!25}70.60$_{\pm 2.58}$\\
         GAT (\textcolor{green}{$\uparrow$ 0.05}) & 73.25$_{\pm11.83}$ & \cellcolor{green!25}74.75$^*_{\pm2.89}$ & \cellcolor{green!25}52.17$_{\pm6.99}$ & 72.06$_{\pm4.81}$ & 60.35$_{\pm 5.06}$ & 70.70$_{\pm3.35}$\\
         GIN (\textcolor{red}{$\downarrow$ 9.70}) & 63.63$_{\pm21.62}$ & 63.60$_{\pm3.88}$ & 52.50$_{\pm5.18}$ & 68.85$_{\pm2.98}$ & 52.63$_{\pm 12.55}$ & 64.30$_{\pm 5.81}$\\\bottomrule
    \end{tabular}
    \label{tab:append_sup}
\end{table}

\end{document}